%% file: main.tex
\documentclass[10pt,twocolumn,letterpaper]{article}

\usepackage{cvpr}              
\input{preamble}
\definecolor{cvprblue}{rgb}{0.21,0.49,0.74}
\usepackage[pagebackref,breaklinks,colorlinks,allcolors=cvprblue]{hyperref}

\usepackage{graphicx}
\usepackage{booktabs}
\usepackage{amsmath,amssymb,amsfonts}
\usepackage{makecell}
\usepackage{bm}
\usepackage{multirow}
\usepackage{graphicx}
\usepackage{amssymb}
\usepackage[accsupp]{axessibility}

\newcommand{\R}{\mathbb{R}}
\newcommand{\DCT}{\text{DCT}_{2D}}

\newcommand{\Conv}[1]{\text{Conv}_{#1\times#1}}
\newcommand{\AvgPool}{\text{AvgPool}}
\newcommand{\MaxPool}{\text{MaxPool}}
\newcommand{\MLP}{\text{MLP}}
\newcommand{\ReLU}{\text{ReLU}}
\newcommand{\BatchNorm}{\text{BatchNorm}}
\newcommand{\Softmax}{\sigma}
\newcommand{\Hadamard}{\odot}

\def\paperID{*****} 
\def\confName{CVPR}
\def\confYear{2026}

\title{AWPD: Frequency Shield Network for Agnostic Watermark Presence Detection}

\author{
    Xiang Ao\textsuperscript{2$\star$}\thanks{$\star$ Equal contribution. This work is accepted by the SAFE Workshop in conjunction with CVPR 2026.} \quad
    Yilin Du\textsuperscript{3$\star$} \quad
    Zidan Wang\textsuperscript{4} \quad
    Mengru Chen\textsuperscript{2} \quad
    Siyang Lu\textsuperscript{1$\dagger$}\thanks{$\dagger$ Corresponding author.}
    \\
    \textsuperscript{1} School of Computer and Information Technology, Beijing Jiaotong University, Beijing, 100044, China\\
    \textsuperscript{2} School of Software Engineering, Beijing Jiaotong University, Beijing, 100044, China\\
    \textsuperscript{3} School of Traffic and Transportation, Beijing Jiaotong University, Beijing, 100044, China\\
    \textsuperscript{4} Law School, Beijing Jiaotong University, Beijing, 100044, China\\
    {\tt\small sylu@bjtu.edu.cn}
}

\begin{document}
\maketitle
\input{sec/0_abstract} 
   
\input{sec/1_intro}
\input{sec/2_related}
\input{sec/AWPD}

\input{sec/FSNet}
\input{sec/Exp}
\input{sec/Conc.tex}
\input{sec/ack.tex}
{
    \small
    \bibliographystyle{ieeenat_fullname}
    \bibliography{main}
}

\clearpage
\appendix
\input{sec/app}

\end{document}

%% file: sec/0_abstract.tex
\begin{abstract}
Invisible watermarks, as an essential technology for image copyright protection, have been widely deployed with the rapid development of social media and AIGC. However, existing invisible watermark detection heavily relies on prior knowledge of specific algorithms, leading to limited detection capabilities for ``unknown watermarks'' in open environments. To this end, we propose a novel task named Agnostic Watermark Presence Detection (AWPD), which aims to identify whether an image carries a copyright mark without requiring decoding information. We construct the UniFreq-100K dataset, comprising large-scale samples across various invisible watermark embedding algorithms. Furthermore, we propose the Frequency Shield Network (FSNet). This model deploys an Adaptive Spectral Perception Module (ASPM) in the shallow layers, utilizing learnable frequency gating to dynamically amplify high-frequency watermark signals while suppressing low-frequency semantics. In the deep layers, the network introduces Dynamic Multi-Spectral Attention (DMSA) combined with tri-stream extremum pooling to deeply mine watermark energy anomalies, forcing the model to precisely focus on sensitive frequency bands. Extensive experiments demonstrate that FSNet exhibits superior zero-shot detection capabilities on the AWPD task, outperforming existing baseline models. Code and datasets will be released upon acceptance.
\end{abstract}

%% file: sec/1_intro.tex
\section{Introduction}
\label{sec:intro}
The explosive development of Generative AI (AIGC)~\cite{sd1,sd2,sd3} and short-video social media platforms has democratized image creation and dissemination, yet this accessibility has simultaneously precipitated an unprecedented crisis in intellectual property (IP) protection and copyright compliance. In this context, invisible watermarking technology has become an essential means of copyright protection due to its advantages of minimal image degradation and high resistance to tampering. With the escalation of application demands, invisible watermarking technology has undergone rapid iteration, evolving from early Least Significant Bit (LSB) steganography~\cite{van1994digital} and Discrete Cosine Transform (DCT) frequency embedding~\cite{cox1997secure, xia1998watermark} to recent deep learning-based~\cite{zhu2018hidden, tancik2020stegastamp} and generative model watermarks~\cite{fernandez2023stable, wen2023tree, deepmind2023synthid}. However, this technological diversity has also spawned a severe problem:

\textbf{The extraction of existing invisible watermarks relies on decoding algorithms that strictly correspond to their embedding methods, making ``agnostic invisible watermark decoding'' almost impossible in practical applications.}

This situation leads image users and platforms into a massive copyright compliance dilemma. Without specific decoders, they cannot determine whether images from unknown sources are copyright-protected, thereby facing an extremely high risk of infringement.

Nevertheless, although agnostic invisible watermark decoding is difficult to achieve, certain commonalities exist among invisible watermarking technologies; namely, to ensure visual imperceptibility, invisible watermarks typically embed features implicitly into the high-frequency bands of images~\cite{barni2001improved, baluja2017hiding, corvi2023intriguing}. Based on this commonality, we attempt to construct a agnostic invisible watermark presence detection(AWPD) mechanism, serving as a Zero-Trust preliminary screening barrier to avoid infringement by identifying the existence of invisible watermarks in images. 

To address such issues, we propose a novel computer vision task: Agnostic Watermark Presence Detection (AWPD). Since it is impossible to ensure that the dataset or model covers all invisible watermarking technologies, the AWPD task places greater emphasis on the detection capability for unknown types of watermarks, that is, the zero-shot performance.

The training of agnostic models heavily relies on data diversity. Therefore, we construct a large-scale comprehensive dataset named UniFreq-100K for the AWPD task. This dataset not only contains watermarked images covering multiple image creation domains and various invisible watermark embedding algorithms, but also provides an equal scale of clean images, offering a benchmark tailored to real-world scenarios for the AWPD task.

Although existing Vision Foundation Models (VFMs) perform exceptionally well in numerous visual tasks~\cite{he2016deep, dosovitskiy2020image, liu2021swin, oquab2023dinov2,yolo,younet}, they tend to capture global semantics and low-frequency spaces~\cite{yin2019fourier} during training to achieve tasks such as classification, detection, and segmentation of visible objects. This characteristic makes them highly prone to treating watermark signals hidden in local and high-frequency bands as noise and discarding them during the downsampling process.

To resolve this dilemma, we propose a novel spatial-frequency joint perception architecture named the Frequency Shield Network (FSNet). Following a Dual-Stage spectral perception paradigm, this model abandons isolated frequency-domain preprocessing steps and deeply integrates the capture of spectral anomalies into different levels of feature extraction, achieving the recognition of invisible watermarks hidden in high frequencies.

Specifically, in the shallowest layer of the network, we deploy the Adaptive Spectral Perception Module (ASPM). This module utilizes the 2D Discrete Cosine Transform (2D DCT) to extract spectral representations and introduces learnable frequency gating to dynamically amplify the extremely weak high-frequency watermark signals while suppressing the dominant low-frequency semantic content. Consequently, it provides robust feature coding for the minute frequency perturbations induced by invisible watermarks before any deep spatial blurring occurs.In the deep stages of the network, to address the issue of information aliasing, we further introduce the Dynamic Multi-Spectral Attention (DMSA) mechanism. This mechanism utilizes multi-branch DCT to recalibrate channel weights and combines tri-stream extremum pooling to deeply mine anomalies in watermark frequency energy, guiding the model to precisely focus on sensitive frequency bands.

Extensive experiments and ablation analyses demonstrate that FSNet significantly outperforms existing detection baselines on the AWPD task, exhibiting superior zero-shot detection capabilities and strong robustness against unseen watermarks.In summary, the main contributions of this paper are as follows:

\begin{itemize}
\item \textbf{Defining a new task and benchmark:} We propose the AWPD task for the first time and construct UniFreq-100K, a large-scale dataset covering multiple watermark generation paradigms, providing a novel research paradigm for digital image copyright protection.

\item \textbf{Proposing a task-specific model baseline:} We deeply analyze the deficiencies of traditional vision models in high-frequency perception and propose FSNet, a novel network architecture combining the ASPM mechanism and DMSA, which effectively compensates for the model's lack of capability in capturing high-frequency microscopic invisible watermarks.

\item \textbf{Validating the model's effectiveness on the task:} We evaluate the performance of mainstream vision models and FSNet on the AWPD task. Among them, FSNet demonstrates performance surpassing baseline models, establishing a new baseline for the AWPD task.
\end{itemize}

%% file: sec/2_related.tex
\section{Related Work}
\subsection{Invisible Watermarks}

Invisible watermarking technology aims to covertly embed information without compromising visual quality. Early traditional methods primarily fall into two categories: spatial domain techniques, which directly modify pixel values (e.g., LSB substitution~\cite{van1994digital} and Patchwork~\cite{bender1996techniques}), and frequency domain techniques, which embed signals into transform coefficients (e.g., DCT~\cite{cox1997secure} or DWT~\cite{xia1998watermark}) to enhance robustness against attacks like JPEG compression.

With the advancement of deep learning, end-to-end neural watermarking frameworks utilizing encoder-decoder architectures (e.g., HiDDeN~\cite{zhu2018hidden}) have become mainstream, significantly improving embedding capacity and efficiency~\cite{stegastamp,editguard,ominguard}. Recently, the explosion of Generative AI (AIGC) has catalyzed exclusive, highly heterogeneous watermarking technologies for generative models~\cite{fernandez2023stable,wen2023tree,deepmind2023synthid}, making the construction of a unified decoder practically impossible.Furthermore, sophisticated defense mechanisms increasingly rely on subtle imaging perturbations rather than conventional watermarking. For instance, recent works have demonstrated that injecting imperceptible adversarial noise to perturb attention mechanisms can efficiently fool customized diffusion models~\cite{Xu_2024_CVPR}, which further exacerbates the complexity of detecting artificial traces in AIGC imagery. 

Notably, despite their evolution, all watermarking algorithms exploit the Masking Effect of the Human Visual System (HVS)~\cite{barni2001improved,reddy2010effective}. To preserve image generation quality, both optimization-based invisible watermarks and generative artifacts tend to concentrate their energy within the microscopic textures and high-frequency residuals of images~\cite{baluja2017hiding,corvi2023intriguing}.

\subsection{Vision Foundation Models}

While Vision Foundation Models (e.g., ResNet~\cite{he2016deep}, ViT~\cite{dosovitskiy2020image}) have significantly advanced visual forensics and zero-shot fake image detection~\cite{chai2020what,ojha2023towards,wang2023dire}, they suffer from an inherent ``low-frequency bias''~\cite{yin2019fourier}. During feature extraction and spatial downsampling, these models tend to capture global semantics while discarding local high-frequency details as noise. This bias severely hinders the AWPD task: since invisible watermarks are typically concealed within these microscopic high-frequency structures, traditional models struggle to capture such extremely weak features.

Recent studies have increasingly integrated frequency-aware mechanisms. Research demonstrates that extracting high-frequency anomalies via Discrete Cosine Transform (DCT) effectively exposes deep forgery clues~\cite{wang2020cnn,qian2020thinking}. Furthermore, the indispensability of frequency domain features for capturing tiny, weak signals has been proven across various domains: HS-FPN~\cite{shi2025hsfpn} utilizes DCT high-pass filtering to prevent feature loss in tiny object detection, while MADGNet~\cite{nam2024madgnet} leverages multi-band DCT and attention mechanisms to overcome low-frequency bias in medical image segmentation. Inspired by these insights, we design FSNet to explicitly amplify and capture high-frequency watermark representations.

\subsection{Existing Agnostic Invisible Watermark Detection}

Currently, agnostic Watermark Detection remains a highly challenging frontier. Existing invisible watermark extraction heavily relies on decoders strictly corresponding to the embedding algorithms. Although in the fields of multimedia forensics and steganalysis, methods such as SRM~\cite{fridrich2012rich} and SRNet~\cite{boroumand2018deep} attempt to detect image steganography by extracting high-frequency noise residuals, these methods are primarily designed for traditional extremely low-capacity steganography.

Faced with modern, complex, and diverse deep learning watermarks and generative watermarks (such as algorithms based on latent space perturbations), traditional steganalysis methods often exhibit severe feature drift and are difficult to generalize across algorithms. Due to the lack of large-scale benchmark datasets covering diverse modern watermarks, existing detection methods possess almost no zero-shot detection capability when facing unseen algorithms.

%% file: sec/AWPD.tex
\begin{figure*}[tb]
\centering
\includegraphics[width=0.75\textwidth]{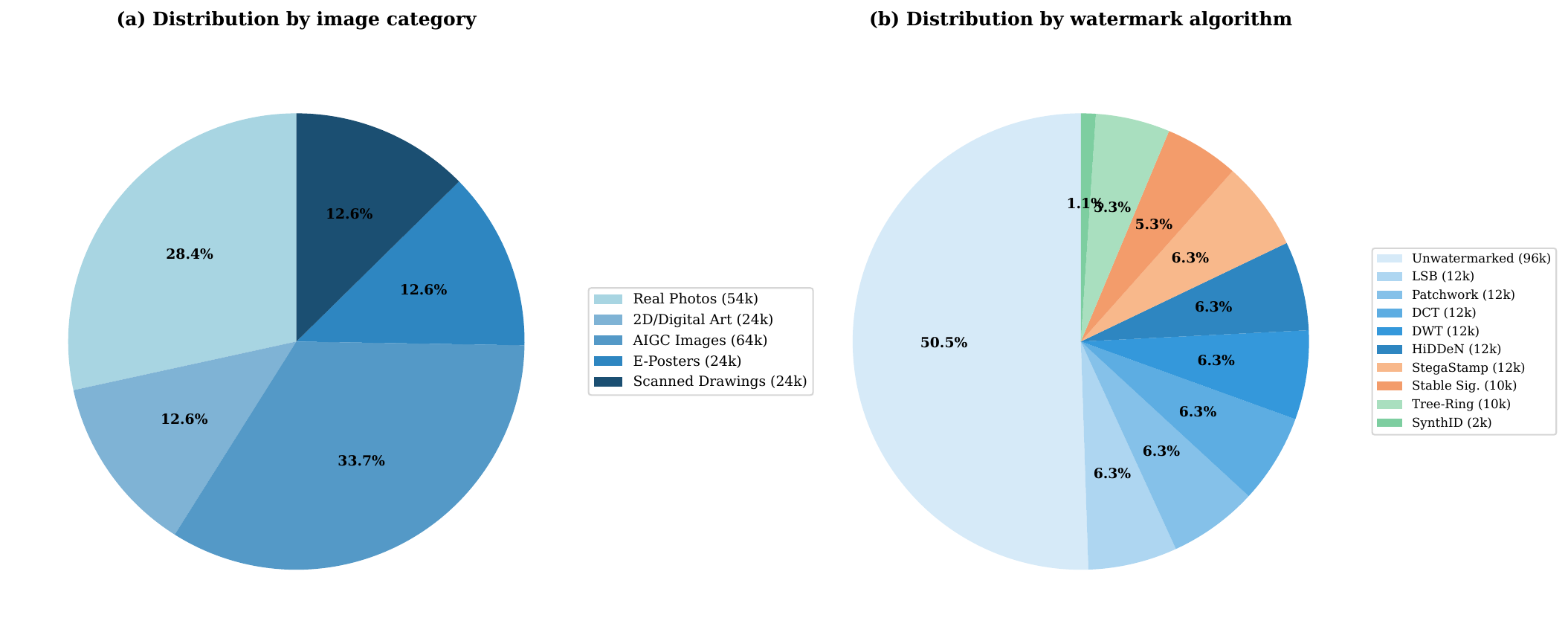}
\caption{Overall distribution of the UniFreq-100K dataset. (a) Distribution across five image categories totaling 190K images. (b) Distribution across watermarking algorithms, including 96K unwatermarked and 94K watermarked images. Evaluated watermarking algorithms: LSB~\cite{van1994digital}, Patchwork~\cite{bender1996techniques}, DCT~\cite{cox1997secure}, DWT~\cite{xia1998watermark}, HiDDeN~\cite{zhu2018hidden}, StegaStamp~\cite{stegastamp}, Stable Sig.~\cite{fernandez2023stable}, Tree-Ring~\cite{wen2023tree}, and SynthID~\cite{deepmind2023synthid}. Detailed cross-distribution data (category $\times$ algorithm) can be found in the Appendix.}
\label{fig:dataset_distribution}
\end{figure*}

\section{Agnostic Watermark Presence Detection}
\label{sec:awpd}

Given an input image $x \in \mathcal{X} \subset \mathbb{R}^{H \times W \times C}$, AWPD aims to learn a binary classification mapping model $\mathcal{F}_{\theta}$ to determine whether an invisible watermark is embedded in the image. We define the label space of this task as $\mathcal{Y} = \{0, 1\}$, where $y=0$ denotes the original clean image and $y=1$ denotes the watermarked image. Therefore, the core objective of AWPD is to optimize the model parameters $\theta$ so that it can accurately perform the mapping determination:
\begin{equation}
\hat{y} = \mathcal{F}_{\theta}(x) \in \{0, 1\}.
\end{equation}

Typically, an image with an invisible watermark $x_{w}$ can be formally represented as:
\begin{equation}
x_{w} = \mathcal{E}_{k}(x_{c}, m), \quad \text{s.t.} \quad \mathcal{D}(x_{w}, x_{c}) \leq \epsilon,
\end{equation}
where $x_{c}$ is the original carrier image, $m$ is the embedded watermark information, $\mathcal{E}_{k}$ is a specific watermark embedding algorithm, $\mathcal{D}(\cdot, \cdot)$ measures the perceptual distance between two images, and the minimal threshold $\epsilon$ ensures visual imperceptibility.

In real-world scenarios, the embedding algorithm may belong to an extremely large and heterogeneous algorithm family $\mathcal{K} = \{k_1, k_2, \ldots, k_N\}$. Since different algorithms will hide the watermark signal in different frequency bands or feature spaces, constructing a agnostic decoder to extract the watermark information is mathematically highly ill-posed. Therefore, AWPD bypasses the complex decoding process and instead focuses on a more agnostic degenerate task: detecting the embedding traces left by any unknown algorithm.

While traditional image classification relies heavily on low-frequency semantics, invisible watermarks preserve macroscopic visual representations to satisfy imperceptibility constraints. Consequently, they conceal perturbation signals within high-frequency artifacts, rendering them imperceptible to the human visual system. This implies that for the AWPD model, there is a need to shift the model from capturing the visible macroscopic features of the image to capturing the invisible high-frequency features.

\section{Experimental Benchmark: UniFreq-100K Dataset}
\label{sec:dataset}

\subsection{Dataset Description}

Considering the rapid iteration and technical heterogeneity of invisible watermarking technologies, it is unrealistic to exhaust all known invisible watermarking algorithms in a single dataset. However, as discussed earlier, to satisfy visual imperceptibility, the vast majority of invisible watermarks tend to embed perturbation signals into the high-frequency bands or complex implicit feature spaces of the images.

Based on this commonality, we employ representative sampling by selecting algorithms from existing mainstream watermarking paradigms, ranging from early spatial and frequency domains to modern deep learning and generative models, to construct the AWPD benchmark dataset—UniFreq-100K.

To ensure the dataset's generalization capability in real-world application scenarios and to accommodate diverse image creation works, UniFreq-100K achieves high diversity in the selection of carrier images. The carrier data not only contains real natural images randomly sampled from the COCO~\cite{coco} dataset but also introduces a massive number of pure digital art images, as well as AIGC images generated by state-of-the-art generative models (such as Gemini 3 and Stable Diffusion) covering multi-dimensional artistic styles.

Meanwhile, the dataset also contains an equal amount of unwatermarked samples covering multiple sources. The ratio of positive to negative samples for each source category is kept close, ensuring sample balance. Ultimately, UniFreq-100K contains 94,000 watermarked images (positive samples) and 96,000 unwatermarked images (negative samples), as illustrated in the distribution charts Figure~\ref{fig:dataset_distribution}.


Details of the dataset construction and copyright information can be found in the Appendix.

\subsection{Evaluation Recommendations}

The core challenge and practical significance of the AWPD task lie in the highly robust recognition of unseen watermarks. If a conventional random split is applied to a mixed dataset, the model is highly prone to overfitting specific known algorithm features, thereby deviating from the original intention of agnostic presence detection.

To this end, we abandon the traditional train/test splitting method and propose using leave-one-algorithm-out cross-validation as the standard evaluation protocol for the AWPD task. 

Specifically, suppose the dataset contains $K$ types of watermarking algorithms in the set $\mathbb{E} = \{e_1, e_2, \dots, e_K\}$ (in this paper $K=9$). In the $i$-th round of cross-validation, we designate the watermarked images generated by a specific algorithm $e_i$ as the unseen test set, while mixing the images generated by the remaining $K-1$ algorithms $\mathbb{E} \setminus \{e_i\}$ to form the training set. By alternately setting different watermarking algorithms as the testing target, we can quantitatively evaluate the model's zero-shot detection capability on unseen watermarking architectures.

Furthermore, to avoid threshold shift caused by class imbalance, during each round of training and testing, we sample unwatermarked images ($y=0$) from the homologous clean image library to strictly match the number of watermarked images at a 1:1 ratio, to form a complete binary classification evaluation data stream.

In addition, the detection robustness of the algorithms can be further tested through methods such as image cropping and compression. Data augmentation can also be applied in the same way during the training phase.

%% file: sec/FSNet.tex
\begin{figure*}[htbp]
\vspace{-0.5cm}
\centering
\includegraphics[width=1\linewidth, trim={0cm 6cm 0cm 3.5cm}, clip]{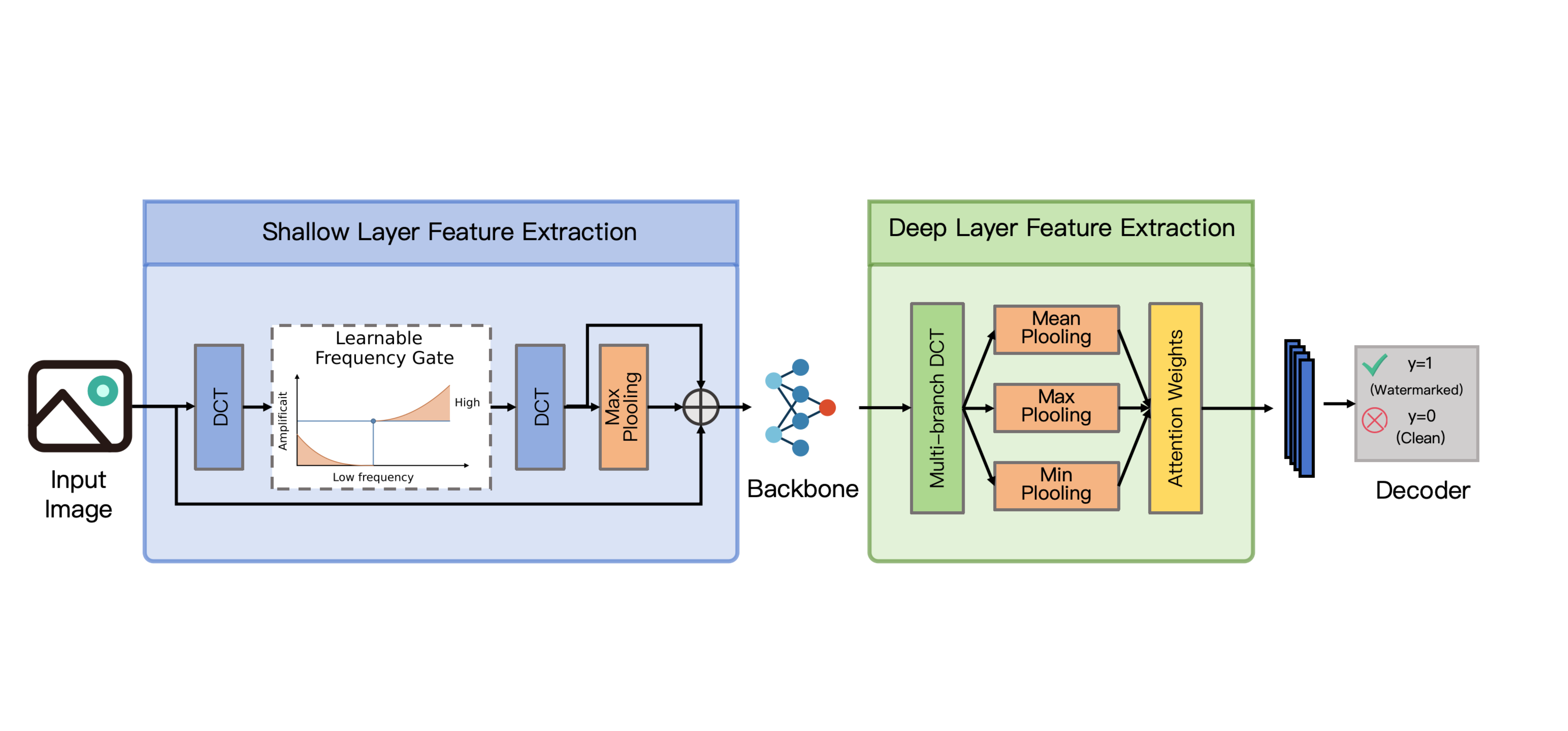}
\caption{Overview of the Frequency Shield Network (FSNet) architecture showing the Adaptive Spectral Perception Module (ASPM) and Dynamic Multi-Spectral Attention (DMSA).}
\label{fig:fsnet}
\end{figure*}

\section{FSNet}
\label{sec:fsnet}

Standard deep learning backbones (e.g., ResNet, ViT) are typically optimized for low-frequency visible information, often treating high-frequency watermark signals as ``noise'' and prematurely discarding them during downsampling. To address the models' deficiency in capturing high-frequency signals, we propose the Frequency Shield Network (FSNet), a spatial-frequency joint perception architecture specifically designed for AWPD, aiming to explicitly capture and retain spectral anomalies. 

As illustrated in Figure~\ref{fig:fsnet}, FSNet follows a Dual-Stage spectral perception paradigm. We do not treat frequency-domain analysis as an isolated preprocessing step, but rather deeply integrate it into the feature extraction hierarchy. This architecture consists of three core components:

\begin{enumerate}
\item \textbf{Adaptive Spectral Perception Module (ASPM):} Deployed at the shallowest layer (Stem) of the network, this module utilizes learnable frequency masks as "soft gating," allowing high-frequency signals related to watermarks to pass while suppressing the dominant low-frequency semantic content.

\item \textbf{Frequency-Aware Backbone:} Employs a standard encoder (we use ResNet-50) to extract deep spatial and semantic features.

\item \textbf{Dynamic Multi-Spectral Attention (DMSA):} Deployed at the deep stages of the network, utilizing multi-branch Discrete Cosine Transform (DCT) to recalibrate channel weights, compelling the network to focus on specific frequency bands where watermark energy is concentrated.
\end{enumerate}

Formally, given an input image $\bm{X} \in \R^{3 \times H \times W}$, the forward encoding process of FSNet can be expressed as:
\begin{equation}
\bm{F}_{stem} = \Phi_{ASPM}(\bm{X}),
\end{equation}
\begin{equation}
\bm{F}_{deep} = \Phi_{Backbone}(\bm{F}_{stem}),
\end{equation}
\begin{equation}
\bm{Z} = \Phi_{DMSA}(\bm{F}_{deep}),
\end{equation}
where $\bm{Z}$ represents the final latent space representation enriched with frequency-domain watermark clues, which is subsequently fed into the classifier $\mathcal{D}$ for authenticity discrimination.

\subsection{Adaptive Spectral Perception Module (ASPM)}

Existing frequency-domain object detection methods typically employ fixed high-pass filters to separate edges and detail information. However, the attacks and defenses of invisible watermarks are extremely diverse. To resist corruptions like JPEG compression, modern watermarking algorithms distribute signals across varying frequencies. Consequently, applying a fixed cutoff frequency introduces a rigid bottleneck, risking the unintended deletion of crucial watermark clues.

To this end, we propose the ASPM, transforming the static filtering process into a learnable dynamic optimization problem. We utilize the 2D Discrete Cosine Transform to convert the input feature map into the frequency domain:
\begin{equation}
\bm{X}_{freq} = \DCT(\bm{X}).
\end{equation}

Unlike traditional binary hard masks, we introduce a learnable spectral gating parameter matrix $\bm{M} \in \R^{H \times W}$. The filtered spectrum $\hat{\bm{X}}_{freq}$ is computed as:
\begin{equation}
\hat{\bm{X}}_{freq} = \bm{X}_{freq} \Hadamard \mathcal{I}(\bm{M}),
\end{equation}
where $\Hadamard$ denotes element-wise multiplication (Hadamard Product), and $\mathcal{I}(\cdot)$ represents bilinear interpolation.

This mask allows all frequencies to pass upon initialization, but during the backpropagation optimization process, it adaptively learns to suppress task-irrelevant frequencies and amplify frequency bands sensitive to watermarks. The filtered spectrum is converted back to the spatial domain via inverse transform to form spatial residuals.

To further amplify the extremely weak watermark signals, we draw inspiration from and improve upon the extremum response mechanism~\cite{shi2025hsfpn}, using adaptive max pooling to capture local extrema, because these extrema often correspond to high-frequency artifacts caused by watermarks in the spatial domain. The final output of ASPM fuses the original input with multi-dimensional residuals:
\begin{equation}
\bm{Y}_{ASPM} = \Conv{3} \Big( \bm{X} + \bm{R}_{spatial} + \MaxPool(\bm{R}_{spatial}) \Big).
\end{equation}

This design ensures that minute frequency perturbations have been sufficiently intercepted and amplified before spatial blurring induced by deep convolutions or splitting occurs.

\subsection{Dynamic Multi-Spectral Attention (DMSA)}

After downsampling and semantic extraction by the backbone network, deep features often suffer from information aliasing, requiring recalibration along the channel dimension (channel-wise) to distinguish watermark features from normal image textures.
Inspired by multi-band analysis~\cite{nam2024madgnet}, we design the DMSA module. Unlike traditional channel attention mechanisms (such as SE-Block) that heavily rely on global average pooling (GAP), DMSA utilizes DCT basis functions to losslessly compress spatial information into spectral descriptors of specific frequencies.

\textbf{Multi-Branch Frequency Decomposition:} We predefine a frequency component index set of size $K$ representing different DCT basis functions. For any specific frequency component $(u, v)$, its DCT basis function $B_{h,w}^{u,v}$ is defined as:
\begin{equation}
B_{h,w}^{u,v} = \cos\left(\frac{\pi h}{H_b}(u + 0.5)\right) \cos\left(\frac{\pi w}{W_b}(v + 0.5)\right).
\end{equation}
To address the issue of deep-layer resolution changes in large models (such as ViT), we avoid forcibly pooling the input features to a fixed $7 \times 7$ size, unlike conventional approaches. Instead, we dynamically interpolate and align the generated basis function weights, thereby maximally preserving the high-frequency fragments in deep feature maps.

\textbf{Tri-Stream Extremum Pooling Mechanism:} Modern invisible watermarks often employ local embedding to reduce visual degradation. The presence of watermarks manifests not only as "peaks" of frequency energy but often also as abnormal "valleys" of local frequency energy (especially in some pixel-subtraction-based embedding algorithms). Traditional average pooling or max pooling can easily miss these covert signals. Therefore, we introduce a tri-stream extremum pooling branch in DMSA.

For the specific spectral feature map $\bm{S}_c$ extracted on channel $c$, we compute three statistical features and fuse them into the final descriptor vector $\bm{v} \in \R^C$:
\begin{equation}
\begin{aligned}
\bm{v}_{avg} &= \AvgPool(\bm{S}), \\
\bm{v}_{max} &= \MaxPool(\bm{S}), \\
\bm{v}_{min} &= -\MaxPool(-\bm{S}),
\end{aligned}
\end{equation}
where the minimum pooling is implemented by taking the negative and then performing max pooling. The final channel attention weight is aggregated through a multi-layer perceptron (MLP):
\begin{equation}
\bm{v}_{total} = \Softmax\left( \MLP \Big( \frac{\bm{v}_{avg} + \bm{v}_{max} + \bm{v}_{min}}{K} \Big) \right).
\end{equation}

This weight $\bm{v}_{total}$ is used for channel re-weighting of deep features. This mechanism endows the model with "frequency band screening" capabilities, enabling it to dynamically suppress channels dominated by low-frequency semantics and assign higher activation values to channels harboring watermark spectra.

\subsection{Frequency Shield Decoder}

To ensure high decoupling of the network architecture and agnostic applicability to different backbones (such as CNNs and Transformers), our decoder adopts a lightweight design. The deep 2D feature map $\bm{Z} \in \R^{C \times H' \times W'}$ outputted by DMSA is compressed into a 1D vector $\bm{z} \in \R^C$ through global adaptive average pooling. Subsequently, it is mapped to binary classification logits through a two-layer MLP with Dropout:
\begin{equation}
\hat{\bm{y}} = \bm{W}_2 \big( \ReLU ( \BatchNorm(\bm{W}_1 \bm{z}) ) \big),
\end{equation}
where $\ReLU$ is the ReLU activation function, and $\bm{W}_1, \bm{W}_2$ are the fully connected layer weights. Since the complex frequency-domain feature mining is handled by the front-end ASPM and DMSA modules, this streamlined classification head effectively prevents the model from overfitting on extremely weak watermark features.

%% file: sec/Exp.tex
\begin{table*}[tb]
\centering
\caption{Comparison of different models on various watermarking and steganography methods (Accuracy and F1-score). \textbf{Bold}: best performance; \textit{Italic}: second-best performance.}
\label{tab:model_comparison}
\resizebox{\textwidth}{!}{
\begin{tabular}{lcccccccccccccc}
\toprule
& \multicolumn{2}{c}{\textbf{FSNet (Ours)}} & \multicolumn{2}{c}{\textbf{CNN}} & \multicolumn{2}{c}{\textbf{ResNet}} & \multicolumn{2}{c}{\textbf{ViT}} & \multicolumn{2}{c}{\textbf{Swin-T}} & \multicolumn{2}{c}{\textbf{DINOv2}} & \multicolumn{2}{c}{\textbf{ConvNeXt-V2}} \\
\cmidrule(lr){2-3} \cmidrule(lr){4-5} \cmidrule(lr){6-7} \cmidrule(lr){8-9} \cmidrule(lr){10-11} \cmidrule(lr){12-13} \cmidrule(lr){14-15}
Method & Acc & F1 & Acc & F1 & Acc & F1 & Acc & F1 & Acc & F1 & Acc & F1 & Acc & F1 \\
\midrule
LSB & \textbf{0.594} & \textit{0.268} & \textit{0.545} & \textbf{0.584} & 0.549 & 0.233 & 0.516 & 0.160 & 0.505 & 0.037 & 0.504 & 0.082 & 0.501 & 0.028 \\
Patchwork & \textbf{0.587} & \textit{0.301} & \textit{0.561} & \textbf{0.641} & 0.544 & 0.284 & 0.519 & 0.150 & 0.529 & 0.207 & 0.506 & 0.106 & 0.500 & 0.009 \\
DWT & \textbf{0.945} & \textbf{0.951} & 0.682 & 0.744 & 0.721 & 0.712 & 0.545 & 0.264 & \textit{0.915} & \textit{0.913} & 0.579 & 0.447 & 0.769 & 0.794 \\
DCT & 0.931 & 0.926 & 0.753 & 0.792 & \textit{0.979} & \textit{0.979} & \textbf{0.991} & \textbf{0.991} & 0.982 & 0.982 & 0.936 & 0.938 & 0.949 & 0.947 \\
HiDDeN & \textbf{0.985} & \textbf{0.989} & 0.829 & 0.843 & \textit{0.977} & \textit{0.977} & 0.950 & 0.948 & 0.964 & 0.965 & 0.882 & 0.886 & 0.889 & 0.891 \\
StegaStamp & \textbf{0.915} & \textbf{0.899} & 0.603 & 0.667 & \textit{0.828} & \textit{0.804} & 0.820 & 0.785 & 0.885 & 0.881 & 0.776 & 0.764 & 0.797 & 0.812 \\
Stable Sig. & 0.901 & 0.925 & 0.667 & 0.622 & 0.935 & 0.938 & \textbf{0.966} & \textbf{0.965} & 0.946 & 0.948 & \textit{0.965} & 0.932 & 0.942 & \textit{0.945} \\
Tree-Ring & \textbf{0.859} & \textbf{0.847} & 0.687 & 0.662 & 0.764 & 0.713 & 0.712 & 0.698 & 0.794 & \textit{0.812} & \textit{0.812} & 0.804 & 0.769 & 0.724 \\
SynthID & \textbf{0.894} & \textbf{0.886} & 0.668 & 0.645 & 0.845 & 0.812 & 0.847 & \textit{0.819} & 0.811 & 0.798 & 0.834 & 0.812 & \textit{0.866} & 0.825 \\
\bottomrule
\end{tabular}
}
\end{table*}

\section{Experiments}
\label{sec:experiments}

In the experimental phase, we evaluated various baseline models as well as our proposed FSNet on UniFreq-100K, thereby verifying the validity of the dataset and the effectiveness of the FSNet architecture, and further providing baseline standards for the AWPD task. Meanwhile, we also conducted additional experiments to further explore the AWPD task and related models. All experiments adopt Cross-Entropy as the loss function and use the AdamW~\cite{adamw} optimizer for training. The learning rate is searched within the range \{0.0001, 0.0005, 0.001, 0.005, 0.01\}, and the batch size is fixed at 128. All experiments are completed on two NVIDIA A100 GPUs. We adopt Accuracy and F1-score as the core evaluation metrics.

Besides FSNet, we also conducted comprehensive evaluations on a series of mainstream and representative baseline models. In terms of classical models, we selected the basic CNN~\cite{cnn} that provides a performance lower bound, as well as ResNet50~\cite{he2016deep}, which serves as the cornerstone of deep convolutional networks. In the attention mechanism paradigm, we evaluated the pioneering ViT-base~\cite{dosovitskiy2020image} and the advanced architecture Swin Transformer-base~\cite{swin_transformer} with efficient hierarchical feature extraction capabilities. Furthermore, to examine the potential of cutting-edge models in invisible watermark recognition, we also introduced DiNO-v2-base~\cite{dinov2}, a self-supervised learning model with strong generalization capabilities, as well as ConvNeXt V2-base~\cite{convnext}, which represents the current leading level of pure convolutional architectures. The pre-trained weights of these baseline models are uniformly obtained from open-source frameworks such as Hugging Face, Torchvision, or timm, thereby ensuring the reproducibility of the experiments.

\subsection{Model Evaluation}

The UniFreq-100K dataset contains 9 types of invisible watermarks. To evaluate the models' generalization ability to recognize unknown watermarks on the AWPD task, we adopted a "Leave-One-Out" cross-evaluation strategy: during the training phase, the model learns the common features of 8 types of watermarks, and during the testing phase, it is used to identify the 9th unseen watermark. In terms of dataset splitting, for each evaluation, we combine the single type of watermark images serving as the test target with an equal number of unwatermarked images to form the test set, and combine the remaining 8 types of watermark images with an equal number of unwatermarked images to form the training set, thereby ensuring that the ratio of positive to negative samples is strictly maintained. The cross-evaluation results of each model on the 9 watermark types are detailed in Table~\ref{tab:model_comparison}.

\begin{table*}[tb]
\centering
\caption{Ablation study results on FSNet components. We evaluate the impact of different modules including the Learnable Gate, Multi-dim Fusion, Multi-freq Branch, and Extremum Pooling.}
\label{tab:ablation}
\small 
\setlength{\tabcolsep}{4pt} 
\resizebox{0.8\textwidth}{!}{
\begin{tabular}{l cc cc cc cc cc cc}
\toprule
\multirow{2}{*}{\textbf{Method}} & \multicolumn{2}{c}{\textbf{Full FSNet}} & \multicolumn{2}{c}{\textbf{w/o Gate}} & \multicolumn{2}{c}{\textbf{w/o Fusion}} & \multicolumn{2}{c}{\textbf{w/o Branch}} & \multicolumn{2}{c}{\textbf{w/o Pooling}} & \multicolumn{2}{c}{\textbf{ResNet}} \\
\cmidrule(lr){2-3} \cmidrule(lr){4-5} \cmidrule(lr){6-7} \cmidrule(lr){8-9} \cmidrule(lr){10-11} \cmidrule(lr){12-13}
& Acc & F1 & Acc & F1 & Acc & F1 & Acc & F1 & Acc & F1 & Acc & F1 \\
\midrule
LSB          & 0.594 & 0.268 & 0.556 & 0.275 & 0.564 & 0.278 & 0.514 & 0.225 & 0.526 & 0.241 & 0.549 & 0.233 \\
Patchwork    & 0.587 & 0.301 & 0.564 & 0.298 & 0.549 & 0.268 & 0.511 & 0.209 & 0.552 & 0.254 & 0.544 & 0.284 \\
DWT          & 0.945 & 0.951 & 0.890 & 0.841 & 0.912 & 0.911 & 0.842 & 0.836 & 0.879 & 0.886 & 0.721 & 0.712 \\
DCT          & 0.931 & 0.926 & 0.921 & 0.924 & 0.914 & 0.905 & 0.945 & 0.928 & 0.939 & 0.933 & 0.979 & 0.979 \\
HiDDeN       & 0.985 & 0.989 & 0.945 & 0.932 & 0.972 & 0.951 & 0.941 & 0.954 & 0.968 & 0.951 & 0.977 & 0.977 \\
StegaStamp   & 0.915 & 0.899 & 0.912 & 0.875 & 0.842 & 0.814 & 0.828 & 0.814 & 0.869 & 0.894 & 0.828 & 0.804 \\
Stable Sig.  & 0.901 & 0.925 & 0.899 & 0.895 & 0.854 & 0.812 & 0.912 & 0.945 & 0.894 & 0.882 & 0.935 & 0.938 \\
Tree-Ring    & 0.859 & 0.847 & 0.845 & 0.821 & 0.844 & 0.832 & 0.794 & 0.745 & 0.795 & 0.786 & 0.764 & 0.713 \\
SynthID      & 0.894 & 0.886 & 0.871 & 0.862 & 0.869 & 0.678 & 0.849 & 0.832 & 0.851 & 0.862 & 0.845 & 0.812 \\
\bottomrule
\end{tabular}
} 
\end{table*}

As seen from the experimental results in Table~\ref{tab:model_comparison}, in the vast majority of cases, the accuracy of all models exceeds 60\%, proving that commonalities do indeed exist in invisible watermark embedding technologies, and agnostic recognition models can be constructed to identify the presence of invisible watermarks. Among them, due to its unique high-frequency capturing design, FSNet outperforms baseline models not specifically designed for this task in most cases, further demonstrating that enhancing high-frequency signals in the AWPD task is beneficial for improving model performance.

However, it is noteworthy that on LSB and Patchwork, the recognition accuracies of baseline models and FSNet are both lower than 60\%. With a positive-to-negative sample ratio of 1:1, such accuracy indicates ineffective classification. We attribute this to the fact that these two methods rely on bit-level manipulation rather than directly embedding high-frequency signals. We will further discuss this issue in the Appendix.

\subsection{Dataset Analysis}

\subsubsection{Scale}

\begin{figure}[htbp]
\centering
\includegraphics[width=1\linewidth]{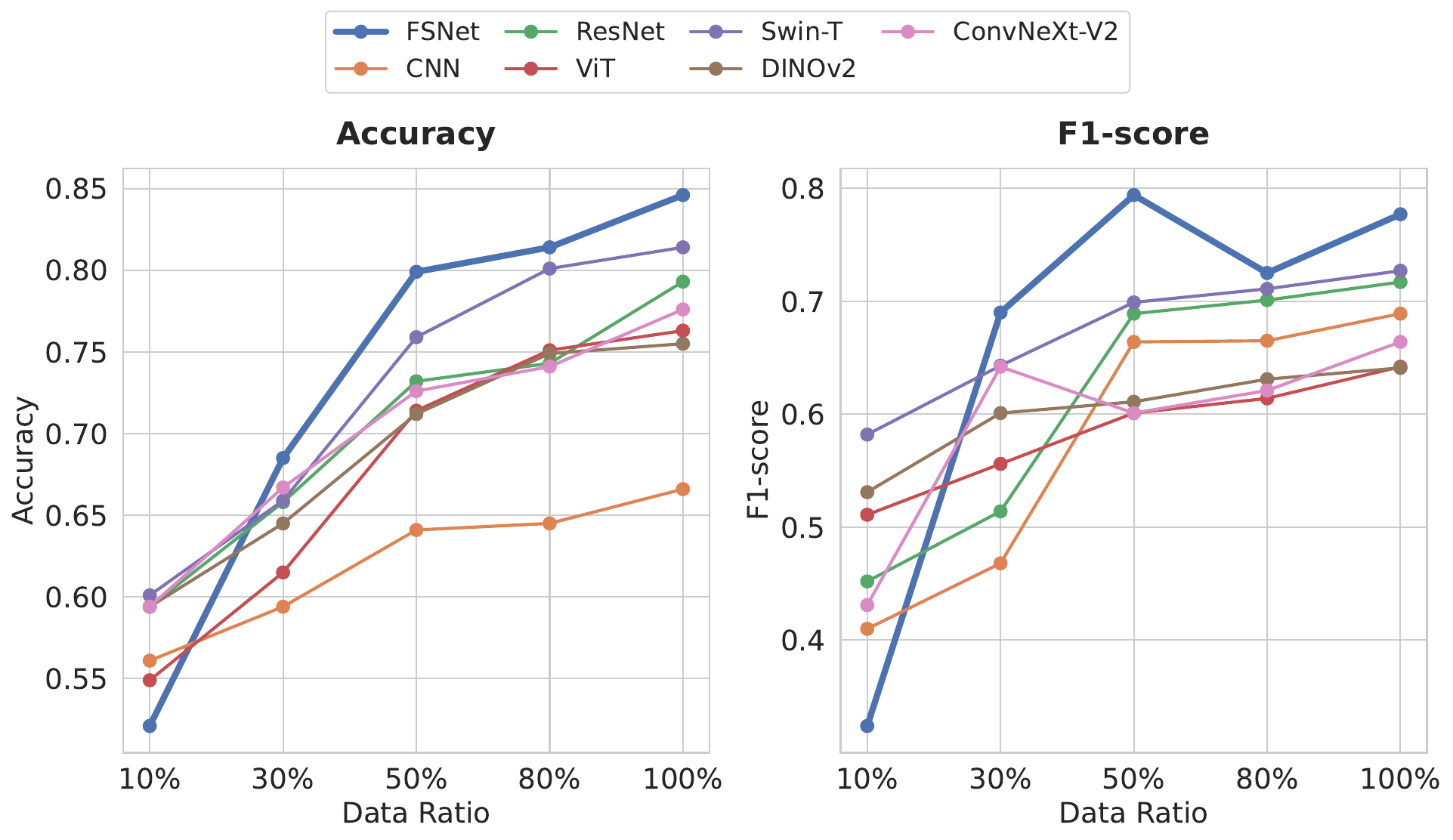}
\caption{Model performance on different dataset ratios (10\% to 100\%). The accuracy increases with dataset size but shows diminishing returns after 30-50\%.}
\label{fig:data_ratio}
\end{figure}

The UniFreq-100K dataset has a relatively large scale, resulting in relatively long training times. Here we adjust different dataset scales to demonstrate the balance between efficiency and accuracy in practical engineering. We tested the accuracy of different models on 10\%, 30\%, 50\%, 80\%, and 100\% of the UniFreq-100K dataset, and plotted a comparison chart, as shown in Figure~\ref{fig:data_ratio}.

It is evident that the dataset scale is positively correlated with accuracy, but marginal effects exist. Under different recognition targets, the sweet spot for performance and efficiency varies, but overall, for the UniFreq-100K dataset, utilizing around 30\% to 50\% of the scale is sufficient.

\subsubsection{Relationships Between Similar Datasets}

\begin{figure}[!htbp]
\centering
\includegraphics[width=1\linewidth]{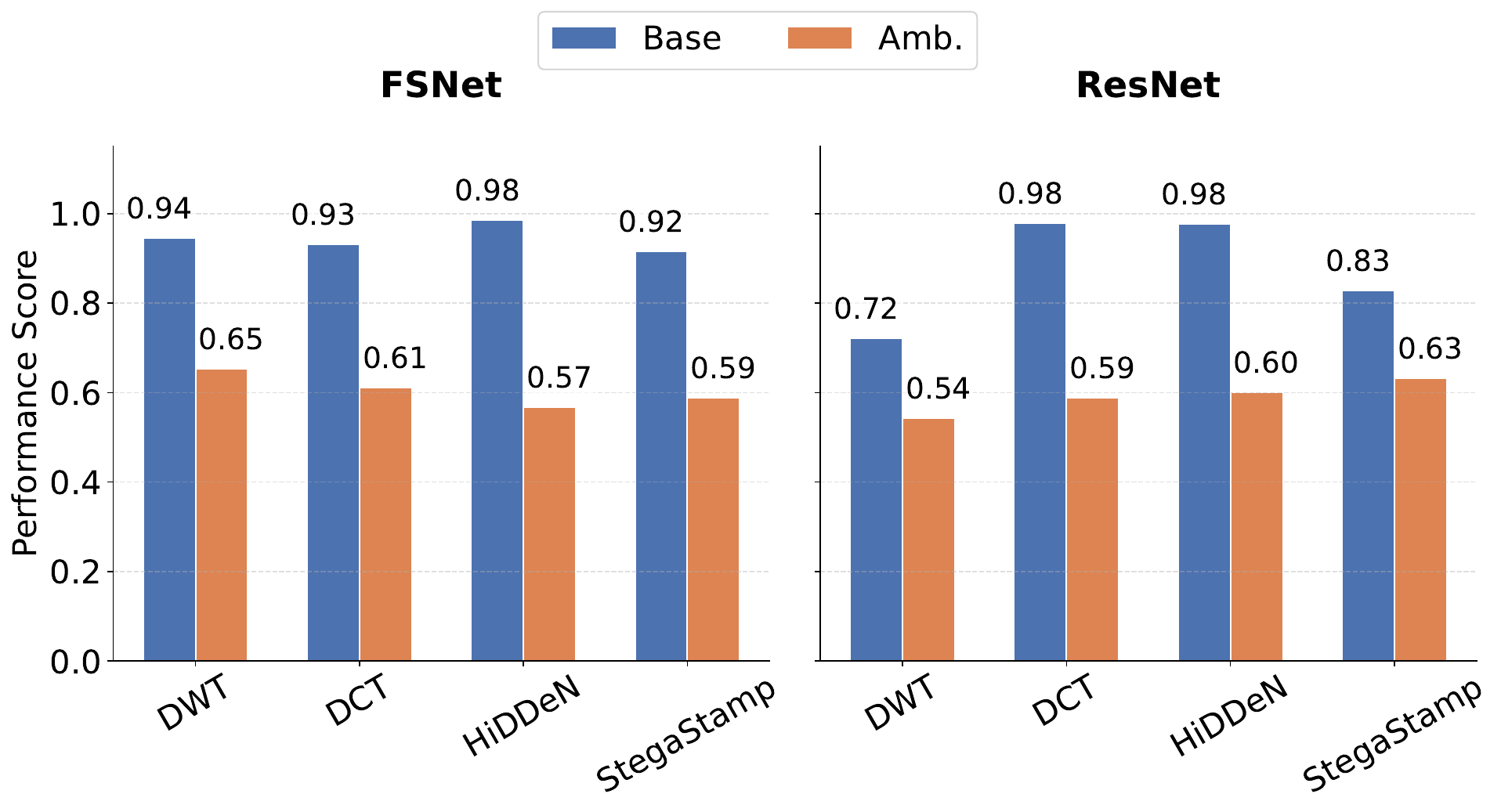}
\caption{Performance comparison after completely removing similar watermark algorithms from the dataset. Base represents the performance without removal, and Amb. represents the performance after removal.}

\label{fig:data_amb}

\end{figure}

Among the 9 watermark types, certain algorithms share similarities, such as DCT and DWT. Here we explore the mutual influence of these similar algorithms. We conducted two experimental schemes: (1) We completely removed the DCT and DWT frequency-domain algorithms, trained on the remaining data, and then evaluated on DCT and DWT datasets. (2) We completely removed the deep learning-based HiDDeN and StegaStamp algorithms, trained on the remaining data, and evaluated on HiDDeN and StegaStamp datasets. The experimental results are shown in Figure~\ref{fig:data_amb}.

The results indicate that model performance declines when similar algorithms are removed, suggesting that shared features among similar algorithms influence recognition. However, FSNet suffered less performance degradation due to its capability of capturing the inherent high-frequency features of invisible watermarks.


\subsection{Ablation Study}

To verify the effectiveness of the FSNet design, we conducted ablation studies with the following configurations: removing the learnable spectral gating in ASPM and replacing it with fixed frequency filtering; removing the multi-dimensional fusion module in ASPM, retaining only the original data and spatial residuals; removing the multi-frequency branches in DMSA and using a unified frequency; removing tri-stream extremum pooling and using average pooling throughout. The experimental results are shown in Table~\ref{tab:ablation}.

The results demonstrate that each component of FSNet contributes to the overall performance. The learnable spectral gating in ASPM allows the model to adaptively focus on watermark-sensitive frequency bands. The multi-dimensional fusion module enhances the representation of high-frequency features. The multi-frequency branches in DMSA enable the model to capture watermark signals across different frequency bands. The tri-stream extremum pooling mechanism effectively captures both energy peaks and valleys caused by watermark embedding.

Furthermore, we conducted additional investigations into FSNet's mechanism for capturing high-frequency signals. Detailed information can be found in the Appendix.

%% file: sec/Conc.tex
\section{Conclusion}
\label{sec:conclusion}

Aiming at the current issue of lacking agnostic detection methods for invisible watermarks, this study proposes the AWPD task and constructs an agnostic invisible watermark detection framework. By constructing a comprehensive dataset covering various invisible watermark algorithms, we provide the essential conditions for agnostic model training. Targeting the common high-frequency features of invisible watermarks, we designed the FSNet model. While FSNet effectively captures the invisible watermark information hidden in high frequencies and achieves optimal performance on our constructed benchmark, we acknowledge that generalizing to entirely unseen and complex real-world environments remains an open challenge. For practical deployment, our method can serve as a preliminary filter in copyright verification systems on social media platforms. Future research will build upon this baseline framework to optimize computational efficiency for large-scale throughput and further explore the detection of watermarks reliant on low-frequency manipulations.

%% file: sec/ack.tex
\section{Acknowledgment}
This research was supported by the General Program of National Natural Science Foundation of China (No.62376023).

%% file: sec/app.tex
\section{UniFreq-100K Dataset Construction Details}
\label{appendix:dataset_details}

To comprehensively evaluate the model's performance on the Agnostic Watermark Presence Detection (AWPD) task, we constructed the large-scale UniFreq-100K dataset. This appendix elaborates on the sources of various watermark carrier images in the dataset, the embedding settings for different watermarking algorithms, specific engineering implementation details, and related copyright statements.

\subsection{Selection and Sampling of Watermarked Images}
To ensure the model can generalize to real-world open scenarios, we sampled from the following distributions when constructing the unwatermarked ``Clean Images'' and the ``Watermarked Images'' used for embedding:
\begin{itemize}
    \item \textbf{Real Photos}: Uniformly and randomly sampled from the widely used COCO 2017 training set (train2017).
    \item \textbf{AIGC Images}: Covering various artistic style images generated by mainstream generative foundation models.
    \item \textbf{2D/Digital Art, Electronic Posters, and Scanned Drawings}: Proportionally collected from various image galleries.
\end{itemize}

\subsection{Watermarking Algorithm and Parameter Settings}

For the 9 representative invisible watermarking algorithms selected, we implemented refined experimental controls during the embedding phase to ensure the fairness and rigor of the evaluation system. Unless otherwise specified, all watermarked images are ultimately uniformly resized to a $256 \times 256$ resolution. To simulate fixed identity tracing in real-world scenarios, we uniformly adopted a random 32-bit binary sequence (32-bit Payload) as the embedding target for traditional algorithms and deep learning baseline algorithms. For algorithms with a payload length exceeding 32 bits, we applied 32-bit cyclic padding or random padding to reach the specified length.

\subsubsection{Traditional Spatial and Frequency Domain Baseline Algorithms}
The specific implementation details for the four classical algorithms (LSB, Patchwork, DCT, DWT) are as follows:

\begin{itemize}
    \item \textbf{Least Significant Bit (LSB) Steganography}: As the basic method for spatial domain steganography, this algorithm directly manipulates the B (Blue) channel in the image's BGR color space. To enhance the watermark's robustness (especially against local cropping attacks), we introduced a full-image redundancy mechanism: the 32-bit target payload is cyclically tiled to cover and replace the least significant bits of all pixels in the B channel of the entire image.
    
    \item \textbf{Patchwork Algorithm}: This is a spatial domain modification technique based on statistical features. For each bit of the 32-bit information, the algorithm uses a fixed random seed to generate 100 pseudo-random pixel pairs. Depending on the embedded bit value (1 or 0), a luminance shift is applied to the corresponding pixel pairs on the B channel, with a fixed modification step size of $d=5$.
    
    \item \textbf{Discrete Cosine Transform (DCT) Watermarking}: The algorithm first converts the image to the YUV color space. A global 2D discrete cosine transform is applied to the Y (luminance) component, concentrating the image energy in the low-frequency region. We use a random seed to generate a standard normal distribution noise pattern and modulate the 32-bit information into this pattern using an intensity coefficient of $\alpha=15.0$. To ensure imperceptibility, the embedding process avoids the direct current (DC) component and only superimposes on the alternating current (AC) coefficients.
    
    \item \textbf{Discrete Wavelet Transform (DWT) Watermarking}: On the Y channel of the YUV space, a two-level decomposition is performed using the Haar wavelet basis, obtaining four frequency sub-bands: LL, HL, LH, and HH. We embed the spread-spectrum random signal carrying the information (intensity coefficient $\alpha=15.0$) into the HL sub-band, which contains horizontal high-frequency information. This method offers strong resistance to cropping and scaling attacks.
\end{itemize}

For the convenience of future reproducibility, the codes for the aforementioned baseline algorithms are provided in Appendix D.

\subsubsection{Deep Learning Watermarking Algorithms}
For invisible watermarks driven by end-to-end neural networks, this study utilized officially released open-source code implementations.
\begin{itemize}
    \item \textbf{HiDDeN}: Utilizes an adversarial training architecture containing an Encoder, a Decoder, and a Noise Layer. We used the training scripts from the original code to train the model. Subsequently, we embedded a random 32-bit code into the images using the trained model.
    
    \item \textbf{StegaStamp}: A robust algorithm designed against physical-world imaging perturbations. Its encoder embeds information into global image features, demonstrating exceptionally strong resistance to rotation, illumination changes, and physical recapture. We implemented it using its open-source code and weights. Since this algorithm can embed actual text content, our embedded payload was ``Hello\_world''.
\end{itemize}

\subsubsection{Generative and AIGC Watermarking Algorithms}
To address the copyright and authenticity identification challenges brought by generative AI such as diffusion models, the dataset integrates watermarking technologies directly injected into the generation pipeline:
\begin{itemize}
    \item \textbf{Stable Signature}: We adopted the official open-source code, using Stable Diffusion 1.5 as the base model. This method lightly fine-tunes the decoder of the LDM, enabling the generated images to automatically carry a specific binary signature when transitioning from the latent space to the pixel space.
    
    \item \textbf{Tree-Ring}: During the initial latent noise phase of the diffusion model, this algorithm embeds information by applying ring-symmetric constraints on the frequency domain energy distribution. We implemented this algorithm manually. During the experiment, the initial image generation size was set to $400 \times 400$, which was ultimately resized to $256 \times 256$. The experiment used a 32-bit message length and a batch size of 32. The encoder and decoder employed a 4-layer convolutional structure with 64 hidden channels and ReLU activation functions. The model was trained for 300 epochs using the Adam optimizer and a Cosine Annealing learning rate scheduling strategy (initial learning rate of $1 \times 10^{-4}$), with a weight decay of $1 \times 10^{-5}$ set to improve generalization. To strengthen detector performance, a complex augmentation combination was introduced during the training phase, including random cropping ($0.2 \sim 0.25$), Dropout ($0.55 \sim 0.6$), scaling ($0.7 \sim 0.8$), and default-enabled JPEG compression perturbations.
    
    \item \textbf{SynthID}: For SynthID watermark samples, we generated them by calling Google's provided \texttt{imagen-4.0-fast-generate-001} model API. This model natively supports injecting advanced, deep learning-based covert watermarks directly into the generated image pixels, deeply coupling the watermark signal with the image semantics while ensuring extremely high generation quality.
\end{itemize}

\subsection{Detailed Cross-Distribution Table}

Table~\ref{tab:dataset_detailed} presents the complete cross-distribution of the UniFreq-100K dataset, showing the exact number of images for each combination of image category and watermarking algorithm.

\begin{table*}[htbp]
\centering
\caption{Detailed cross-distribution of image categories and watermarking algorithms in the UniFreq-100K dataset (all values in thousands)}
\label{tab:dataset_detailed}
\small
\setlength{\tabcolsep}{4pt}
\begin{tabular}{lcccccc}
\toprule
\textbf{Algorithm} & \textbf{Real} & \textbf{2D/Art} & \textbf{AIGC} & \textbf{E-Post.} & \textbf{Scan.} & \textbf{Total} \\
\midrule
Unwatermarked & 30 & 12 & 30 & 12 & 12 & 96 \\
\midrule
LSB & 4 & 2 & 2 & 2 & 2 & 12 \\
Patchwork & 4 & 2 & 2 & 2 & 2 & 12 \\
DCT & 4 & 2 & 2 & 2 & 2 & 12 \\
DWT & 4 & 2 & 2 & 2 & 2 & 12 \\
HiDDeN & 4 & 2 & 2 & 2 & 2 & 12 \\
StegaStamp & 4 & 2 & 2 & 2 & 2 & 12 \\
Stable Sig. & 0 & 0 & 10 & 0 & 0 & 10 \\
Tree-Ring & 0 & 0 & 10 & 0 & 0 & 10 \\
SynthID & 0 & 0 & 2 & 0 & 0 & 2 \\
\midrule
\textbf{Watermarked Sub.} & \textbf{24} & \textbf{12} & \textbf{34} & \textbf{12} & \textbf{12} & \textbf{94} \\
\midrule
\textbf{Grand Total} & \textbf{54} & \textbf{24} & \textbf{64} & \textbf{24} & \textbf{24} & \textbf{190} \\
\bottomrule
\end{tabular}
\end{table*}

As shown in Table~\ref{tab:dataset_detailed}, the dataset maintains a balanced distribution across different image categories while accommodating the specific characteristics of different watermarking algorithms. Notably, generative watermarking algorithms (Stable Signature, Tree-Ring, and SynthID) are exclusively applied to AIGC images, as these algorithms are specifically designed for integration with generative models.

\subsection{Dataset Copyright and Open-Source Statement}
The UniFreq-100K dataset constructed in this study strictly complies with the open-source licenses and copyright constraints of relevant data and codes to ensure no intellectual property disputes. The specific copyright statements are as follows:
\begin{itemize}
    \item \textbf{Open-Source Algorithm Codes and Models}: The implementations used to generate some deep learning and generative watermark samples (such as HiDDeN, StegaStamp, Stable Signature, etc.) in the dataset are directly used or based on their official open-source repositories. We strictly adhered to the respective open-source licenses of each algorithm during dataset generation, utilizing them solely for non-commercial academic research purposes.
    \item \textbf{Real Photos}: The real natural image carriers proportionally sampled in the dataset are sourced from the public COCO 2017 dataset. The use and redistribution of this data fully comply with the COCO dataset's original terms of use (Creative Commons Attribution 4.0 License).
    \item \textbf{Paid Licensed Images}: Other types of images included in the dataset (such as 2D/digital art, electronic posters, scanned drawings, etc.) were purchased through formal channels by the research team. Additionally, the team purchased access to Google's paid image generation API. We possess full legal usage rights and distribution authorization for academic research purposes for these images, ensuring the compliance of the benchmark test data.
    \item \textbf{Open-Source Commitment}: Due to anonymous submission requirements and file size limitations, we cannot upload the complete dataset and open-source links at this time. The UniFreq-100K dataset will be open-sourced to the academic community following the formal acceptance of this paper. The open-sourcing of the dataset will be accompanied by a Non-Commercial Research License. Users must simultaneously comply with the copyright agreements of the aforementioned original image sources and baseline algorithms when downloading and using the data.
\end{itemize}

\section{Degradation Analysis of Agnostic Invisible Watermark Detection Performance Based on Weak Spatial Characteristics}

\begin{figure*}[htbp]
    \centering
    \begin{subfigure}[b]{0.24\textwidth}
        \centering
        \includegraphics[width=\textwidth]{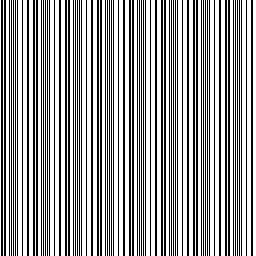}
        \caption{LSB}
    \end{subfigure}
    \hfill
    \begin{subfigure}[b]{0.24\textwidth}
        \centering
        \includegraphics[width=\textwidth]{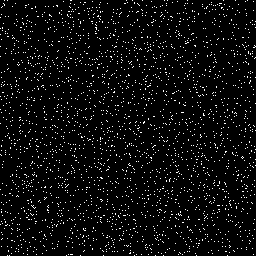}
        \caption{Patchwork}
    \end{subfigure}
    \hfill
    \begin{subfigure}[b]{0.24\textwidth}
        \centering
        \includegraphics[width=\textwidth]{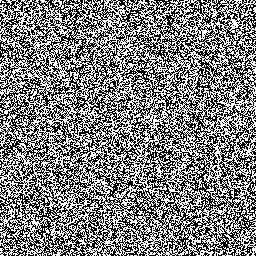}
        \caption{DCT}
    \end{subfigure}
    \hfill
    \begin{subfigure}[b]{0.24\textwidth}
        \centering
        \includegraphics[width=\textwidth]{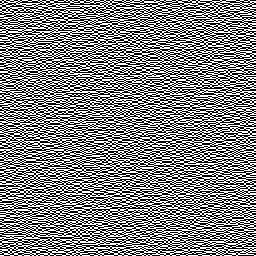}
        \caption{DWT}
    \end{subfigure}
    \caption{Visualization of absolute residual extremum binarization triggered by four watermarking algorithms under a pure white background. From left to right: Dense periodic vertical stripes of LSB; extremely sparse impulse dot matrix of Patchwork; continuous and dense high-frequency grid perturbations of DCT and DWT.}
    \label{fig:residuals}
\end{figure*}

In the evaluation experiments presented in the main text, we observed a significant common phenomenon: all mainstream visual baseline models, encompassing both Convolutional Neural Networks (e.g., ResNet) and Vision Transformers (e.g., ViT), including the FSNet proposed in this paper, exhibited severe performance degradation when detecting the two traditional spatial domain watermarking algorithms: Least Significant Bit (LSB) and Patchwork. To isolate the interference of the complex semantic background noise of natural images and intuitively explore this underlying mechanism, we designed a set of rigorous, controlled visualization experiments. Specifically, we constructed a pure white background image with a resolution of $256 \times 256$ as the carrier and embedded the same 32-bit random payload using the LSB, Patchwork, DCT, and DWT algorithms, respectively. Subsequently, we calculated the absolute residual matrix between the original image and the watermarked image, extracted the maximum variation across channels, and applied extremum visual amplification (i.e., any minute perturbation is binarized to a pure white pixel). The residual visualization results of this experiment are shown in Figure \ref{fig:residuals}. Based on these results, we conducted an in-depth mathematical and physical analysis from the dimensions of spatial sparsity and amplitude signal-to-noise ratio.

\subsection{Patchwork: Extreme Spatial Sparsity and Signal Annihilation During Downsampling}

The core embedding logic of the Patchwork algorithm relies on using a pseudo-random sequence to select a very small number of pixel pairs in the image and applying minute positive and negative shifts to their luminance values. As shown in the Patchwork residual results in Figure \ref{fig:residuals}, the pixel-level modifications it introduces manifest as an extremely sparse array of isolated impulses in 2D space. Assuming the original input image is $x_c \in \mathbb{R}^{H \times W}$, the image after embedding the Patchwork watermark can be represented as $x_w = x_c + \delta_{pw}$, where the number of non-zero elements in the perturbation matrix $\delta_{pw}$ is far less than the total number of pixels in the image, i.e., $\|\delta_{pw}\|_0 \ll H \times W$.

Regardless of the architecture employed by modern visual foundation models, the core of their hierarchical representation construction relies heavily on the spatial downsampling mechanism. Let a local pooling or downsampling operation of a certain layer in the network be $\mathcal{P}(\cdot)$. Within the receptive field $\Omega$, due to the extreme spatial sparsity of $\delta_{pw}$, in the vast majority of local neighborhoods, $\delta_{pw}(i,j) = 0$. Even in neighborhoods containing perturbations, the local pooling operation will cause the sparse high-frequency impulses to be diluted by the surrounding dominant constant pixel values:

\begin{equation}
\mathcal{P}_{\Omega}(x_w) = \mathcal{P}_{\Omega}(x_c + \delta_{pw}) \approx \mathcal{P}_{\Omega}(x_c)
\end{equation}

This extreme sparsity causes the microscopic geometric anomalies triggered by the Patchwork watermark to be irreversibly smoothed or discarded early in the network's forward propagation, rendering general binary classifiers unable to find effective decision boundaries in the deep feature space.

\subsection{LSB: Extremely Low-Amplitude Perturbations and Feature Masking at the Normalization Level}

In stark contrast to the sparse distribution of Patchwork, the LSB algorithm embeds by directly replacing the least significant bit of the pixel channels. As shown on the far left of Figure \ref{fig:residuals}, the modifications brought by cyclically tiling the 32-bit payload information achieve complete spatial alignment, forming extremely dense, periodic vertical stripes. However, the fundamental reason deep models falter on LSB lies not in its spatial distribution, but in its extreme low-amplitude characteristics. The maximum modification amplitude of LSB is strictly confined to an extremely tiny range (i.e., $\pm 1$), and the energy level of this perturbation signal $\delta_{lsb}$ is typically around $1/255$.

In deep neural networks, to ensure training stability, normalization operations (such as Batch Normalization) are widely deployed at every layer. Assuming the feature map input of a certain hidden layer dimension is $X = X_c + \Delta X_{lsb}$, the computation at the normalization layer can be formalized as:

\begin{equation}
\hat{X} = \frac{(X_c + \Delta X_{lsb}) - \mu}{\sqrt{\sigma^2 + \epsilon}} = \frac{X_c - \mu}{\sqrt{\sigma^2 + \epsilon}} + \frac{\Delta X_{lsb}}{\sqrt{\sigma^2 + \epsilon}}
\end{equation}

In natural images, the variance $\sigma^2$ brought by macroscopic semantic content is absolutely dominant, and $\sigma^2 \gg \text{Var}(\Delta X_{lsb})$. This means that after normalization, the relative response value of the extremely low-amplitude perturbations introduced by LSB is severely compressed. Combined with floating-point precision truncation and non-linear activation functions, this ultra-weak signal, which falls below the noise floor of the network, rapidly undergoes ``Feature Masking,'' naturally being treated as meaningless quantization error by the model and consequently filtered out.

\subsection{Accurate Capture of Dense Frequency Anomalies and Agnostic Detection Boundaries}

To further verify the aforementioned mechanisms, we compared the controlled experimental results of the transform domain-based DCT and DWT watermarks. As shown in the two rightmost images of Figure \ref{fig:residuals}, after the inverse transform back to the spatial domain, frequency domain embedding forms interwoven, dense, and continuous ``high-frequency grid perturbations.'' This perturbation is not only absolutely dense spatially (successfully overcoming Patchwork's sparse pooling dilution issue), but due to the inverse mapping diffusion of frequency domain energy, its local microscopic structure possesses sufficient variance to penetrate the network's normalization layers (successfully overcoming LSB's low-amplitude masking issue).

The experiments and derivations above profoundly reveal the capability boundaries of current models in the AWPD task. Existing visual models (including FSNet, which performs specific high-frequency enhancement) inherently rely heavily on capturing continuous frequency anomalies with a certain energy density. LSB and Patchwork successfully bypassed the feature extraction preferences of modern networks by exploiting extreme spatial sparsity and low amplitude. However, because these early algorithms completely lack robustness against routine image post-processing like JPEG compression, they have been eliminated in real-world copyright protection scenarios. Therefore, the models' performance on such algorithms paradoxically proves their accurate locking capability on the core common feature of modern invisible watermarks: ``dense high-frequency spectral anomalies.''

\section{Visual Analysis of the Feature Capture Mechanism of FSNet}
\label{sec:appendix_c}

In the analyses of the main text and Appendix B, we theoretically demonstrated the dilemmas faced by traditional visual baseline models in the AWPD task. To intuitively dissect how FSNet breaks through this limitation and achieves outstanding zero-shot generalization capabilities under extremely weak signals, we designed two sets of in-depth controlled visualization experiments targeting the network's core innovative modules (ASPM and DMSA).

\begin{figure}[htbp]
    \centering
    \includegraphics[width=1\linewidth]{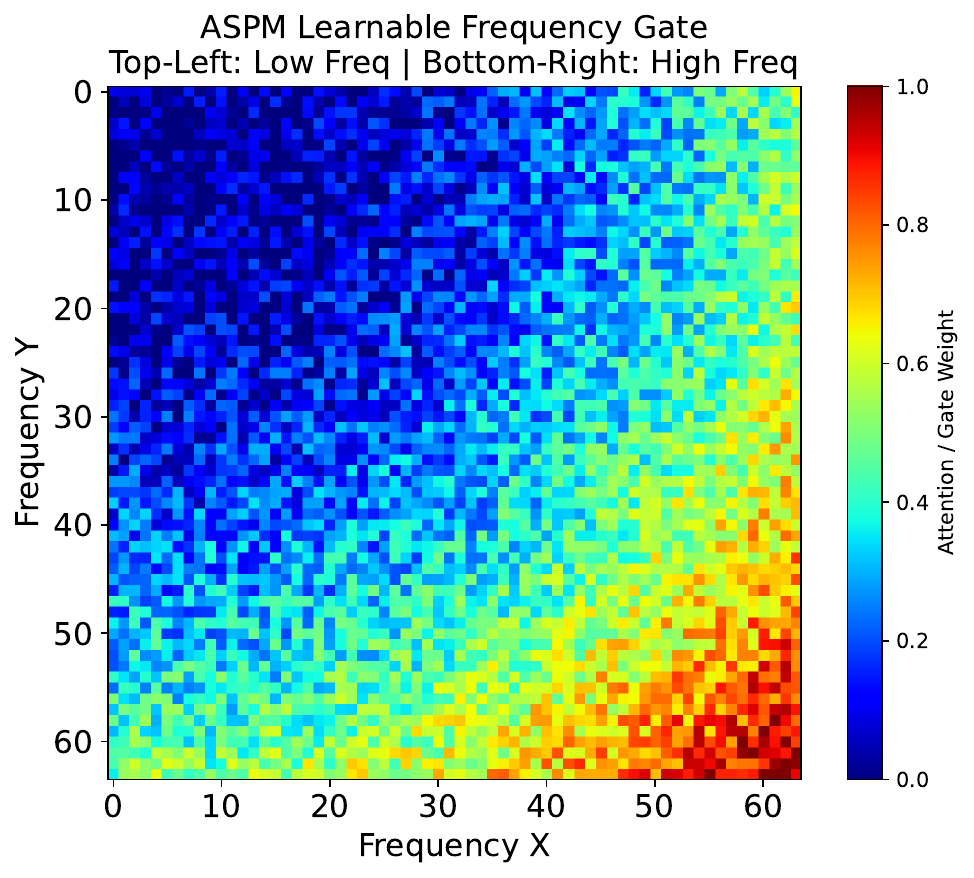}
    \caption{Learnable frequency domain gating heatmap of ASPM after training convergence. The weights in the top-left low-frequency region are significantly suppressed, while the weights in the mid-to-high-frequency regions are adaptively amplified.}
    \label{fig:exp1_aspm}
\end{figure}

\subsection{Shallow Frequency Domain Gating Analysis}
\label{subsec:aspm_vis}

The Adaptive Spectrum Perception Module (ASPM) is deployed at the shallowest layer of the network. Its core is to dynamically intercept and amplify high-frequency watermark signals through a Learnable Frequency Gate. To verify whether this module operates as expected, we extracted the model's weight parameters after training convergence on the UniFreq-100K dataset and mapped them into a pseudo-color heatmap in 2D polar coordinates.

As shown in Figure \ref{fig:exp1_aspm}, this frequency domain mask exhibits a highly regular activation distribution in the 2D spectral space. In the 2D DCT spectrum, the top-left corner typically corresponds to the low-frequency direct current (DC) component of the image, representing macroscopic semantics and smooth backgrounds; whereas the bottom-right and edge regions correspond to high-frequency alternating current (AC) components, containing rich edge textures and minute perturbations.

It can be clearly observed that, driven by backpropagation optimization, the model adaptively assigned extremely low pass-through weights (appearing in cool tones) to the low-frequency region in the top-left, while allocating significantly high weights (appearing in warm tones) to the mid-to-high-frequency regions. This visualization compellingly proves that ASPM successfully acts as a data-driven ``dynamic high-pass filter.'' Before irreversible spatial downsampling occurs in deep networks, it proactively strips away the dominant background noise of natural images, forcing the model to complete a fundamental perspective shift from ``capturing macroscopically visible semantics'' to ``capturing microscopically invisible artifacts.''

\begin{figure}[htbp]
    \centering
    \includegraphics[width=1\linewidth]{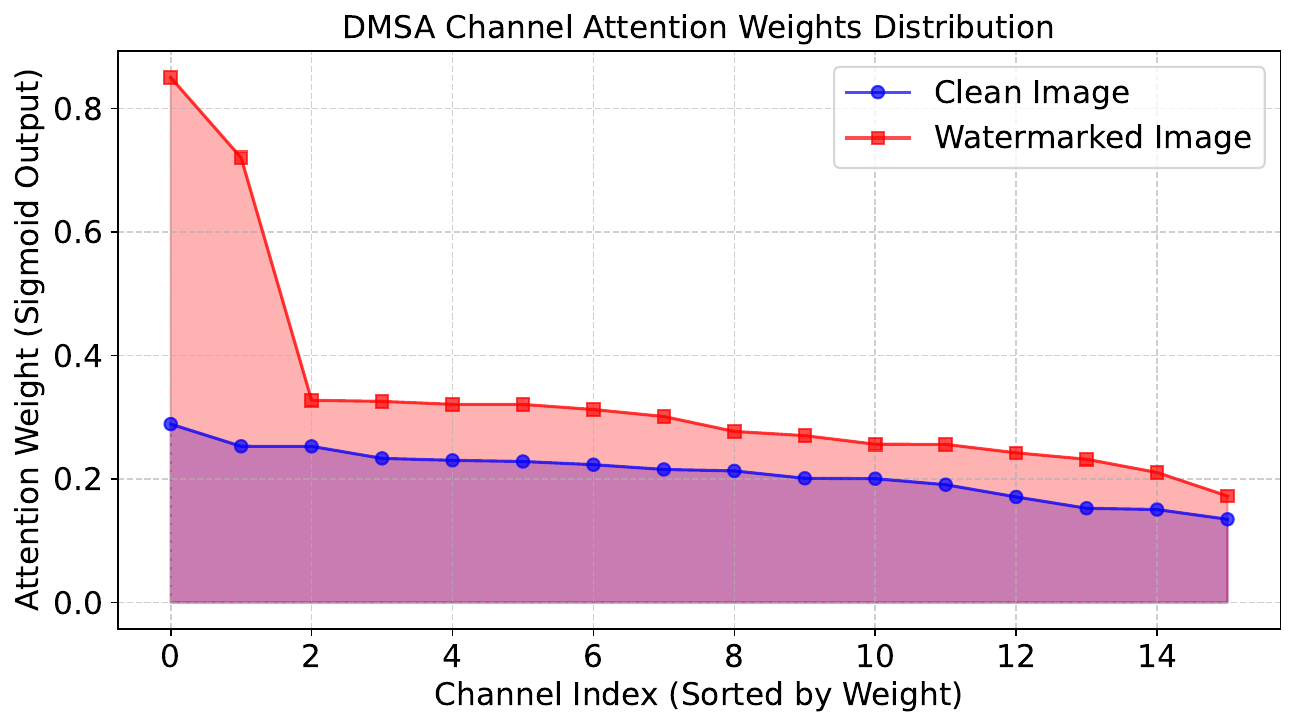}
    \caption{Comparison of channel attention weight distributions of the DMSA module under clean images and watermarked images. Watermarked images trigger significant activation peaks in specific frequency bands.}
    \label{fig:exp2_dmsa}
\end{figure}

\subsection{Deep Multi-band Attention Analysis}
\label{subsec:dmsa_vis}

Retaining high-frequency signals at shallow layers is not sufficient to handle the complex heterogeneity of unknown watermarking algorithms in the AWPD task. Modern invisible watermarks often concentrate energy in specific narrow-band frequencies. The primary design intention of the Dynamic Multi-band Attention mechanism (DMSA) is to recalibrate channel weights in the deep feature space using multi-branch discrete cosine transforms and extremum pooling.

To explore the channel activation preferences of DMSA, we fed Clean Images and Watermarked Images into the network separately and extracted the attention weight distributions of the 16 specific frequency branches output by the DMSA module.

As indicated by the distribution curves in Figure \ref{fig:exp2_dmsa}, when the input is a clean image, the attention weights across all frequency band channels remain in an overall suppressed state with a relatively flat distribution, representing the normal baseline noise floor of the natural image's deep features. However, when the input is an image containing an invisible watermark, the weight curve exhibits dramatic ``impulse-like'' peaks at specific frequency indices.

This significant distributional difference indicates that the DMSA module does not blindly respond to all high-frequency noise. Instead, it possesses an accurate localization capability similar to a ``frequency radar.'' Faced with spectral energy anomalies injected by different concealment algorithms, DMSA can acutely capture local energy fluctuations (whether peaks or valleys) through three-stream extremum pooling. It dynamically assigns extremely high activation weights to the feature channels encompassing these sensitive frequency bands, thereby refining the most critical true-or-false decision boundaries for the classifier within the information-aliased deep space.